\documentclass[runningheads]{llncs}
\usepackage{graphicx}
\usepackage[dvipsnames]{xcolor}

\title{Simultaneous Image Quality Improvement and Artefacts Correction in Accelerated MRI}%
\titlerunning{Simultaneous Quality and Artefacts Correction in Accelerated MRI}

\author{Georgia Kanli\inst{1,2,3} \and Daniele Perlo\inst{1} \and  Selma Boudissa\inst{1,2}\and  Radovan Jiřík\inst{4}  \and  Olivier Keunen\inst{1,2}}

\authorrunning{Kanli G et al.}
\institute{Translational Radiomics, Luxembourg Institute of Health, Luxembourg \and In-Vivo Imaging Platform, Luxembourg Institute of Health, Luxembourg \and Faculty of Electrical Engineering and Communication, Brno University of Technology, Czech Republic \and Institute of Scientific Instruments of the Czech Academy of Sciences, Czech Republic}

\begin{document}
\maketitle

\begin{abstract}
MR data are acquired in the frequency domain, known as k-space. Acquiring high-quality and high-resolution MR images can be time-consuming, posing a significant challenge when multiple sequences providing complementary contrast information are needed or when the patient is unable to remain in the scanner for an extended period of time. Reducing k-space measurements is a strategy to speed up acquisition, but often leads to reduced quality in reconstructed images. Additionally, in real-world MRI, both under-sampled and full-sampled images are prone to artefacts, and correcting these artefacts is crucial for maintaining diagnostic accuracy. Deep learning methods have been proposed to restore image quality from under-sampled data, while others focused on the correction of artefacts that result from the noise or motion. No approach has however been proposed so far that addresses both acceleration and artefacts correction, limiting the performance of these models when these degradation factors occur simultaneously. To address this gap, we present a method for recovering high-quality images from under-sampled data with simultaneously correction for noise and motion artefact called USArt (Under-Sampling and Artifact correction model). Customized for 2D brain anatomical images acquired with Cartesian sampling, USArt employs a dual sub-model approach. The results demonstrate remarkable increase of signal-to-noise ratio (SNR) and contrast in the images restored. Various under-sampling strategies and degradation levels were explored, with the gradient under-sampling strategy yielding the best outcomes. We achieved up to $5\times$ acceleration and simultaneously artefacts correction without significant degradation, showcasing the model's robustness in real-world settings.
\end{abstract}

\keywords{magnetic resonance imaging, acceleration, under-sampling, artefact/noise correction, deep learning}

\section{Introduction}

Magnetic Resonance Imaging (MRI) provides detailed anatomical and functional information on soft tissues by collecting raw data in k-space~\cite{Moratal2008}. The MRI scan time is influenced by the number of phase encoding steps required to reconstruct an image. Increasing the resolution or quality of an image typically requires more phase encoding steps, leading to longer scan times. This poses challenges, especially for patients who struggle to remain still, such as children and people with claustrophobia or uncontrolled movements disorder. Reducing scan time improves patient comfort and lowers medical costs by increasing throughput, but it can also reduce image quality~\cite{Wang2021,Hollingsworth2015,Huang2022}. 

Accelerating MRI acquisition has been a major focus in the field~\cite{Sriram2020,Ramzi2020,Haji2021,Eo2018,Hyun2018,Muckley2021,Zbontar2018,Cheng2019}, leveraging both physics-driven and data-driven strategies to reduce scan times and improve image quality. Physics-driven methods like parallel imaging (SENSE, GRAPPA), compressed sensing (CS), and Echo Planar Imaging (EPI) exploit physical principles to decrease acquisition time but can introduce artefacts like noise amplification, residual aliasing, and Nyquist ghosting. Data-driven approaches, particularly deep learning with convolutional neural networks~\cite{O'Shea2015} (CNNs) or autoencoders, such as U-net~\cite{Ronneberger2015}, have also been proposed to predict missing data resulting from various under-sampled k-space data acquisition strategies, offering robust image reconstruction. 

In real-world MRI, signal acquisition is also subject to various degradation factors that cause artefacts, which are undesired and unreal information that appear in the reconstructed image. Although full sampling may theoretically provide the most complete data, it is still susceptible to motion, noise and other imperfections inherent in the imaging process. These issues distort anatomical structures, introduce false information or cause signal loss, compromising diagnostic accuracy. Therefore, in the pursuit of improving image quality through acceleration techniques, it's crucial to not only address artefacts arising from under-sampling but also to carefully manage and mitigate additional artefacts inherent to imperfect acquisitions settings that may be further amplified by the under-sampling process. 

In the present paper, we present a new method to restore quality in images reconstructed from under-sampled MRI data acquisitions. We propose a neural network model called USArt (Under-Sampling and Artifact correction model) that restores missing k-space data and simultaneously corrects for motion and noise related artefacts. By addressing both under-sampling and artefacts correction, we aim to enhance the overall quality and accuracy of real-life fast MRI methods, thus contributing to more reliable and effective image reconstruction techniques. Our approach involves a dual-model framework, featuring one model operating in the k-space domain and another in the image domain, inspired by prior work that exploits the different characteristics of these domains ~\cite{Haji2021,Eo2018,Cheng2019}. We examined different under-sampling strategies, acceleration factors and artefacts. This project focuses on single-channel coils and Cartesian sampling for simplicity; hence, parallel MRI is not discussed.

\section{Methods}

\subsection{Dataset description}
In vivo 2D T2w anatomical images of mouse brain with tumors were acquired according to established protocols \cite{Oudin2021,Golebiewska2020}.
Images acquisitions used a Cartesian sampling trajectory and were performed on a preclinical 3T MRI system (MRSolutions, UK) equipped with a quadrature head coil. 
The dataset consists of 5649 complex-valued images from 224 different mice. The train/validation dataset includes 5000/449 images from 204/30 subjects. The remaining 200 images from 10 different subjects were used to test the performance of the trained networks. There was no overlapping of the same subjects or images in the different datasets. 

\subsection{Under-sampling}
Under-sampling was achieved by retrospectively dropping lines in fully acquired Cartesian k-spaces, corresponding to phase-encoding steps. Masks were for this purpose applied to selectively zero out lines in the k-space using one of three strategies: gradient, random, and uniform under-sampling.
The gradient under-sampling mask progressively reduces the number of lines acquired as the trajectory moves away from the k-space center. This favors low frequency k-space information that determines contrast, brightness, and general shapes, over high frequencies that pertain to edges and sharp transitions. Random under-sampling selects the retained lines randomly. Uniform under-sampling uniformly discard lines in the k-space, without targeting specific regions or frequencies. Three under-sampling acceleration factors were used: $2\times$, $5\times$, and $10\times$. For all under-sampling strategies, we additionally retained some low-frequency lines in order to reduce the aliasing artefact; 25\%, 10\%, and 4\% of the k-space's lines for $2\times$, $5\times$, and $10\times$ accelerations respectively.

\subsection{Artefacts}
Artefacts in real-world acquisitions were simulated in the k-space domain using a custom developed library\cite{Boudissa2024}, accessible through GitHub\footnote{https://github.com/TransRad/MRArt.git}. We explored the most common artefacts in anatomical images: Gaussian noise and motion artefact.

\begin{figure}[h]
  \centering
  \includegraphics[width=1\textwidth]{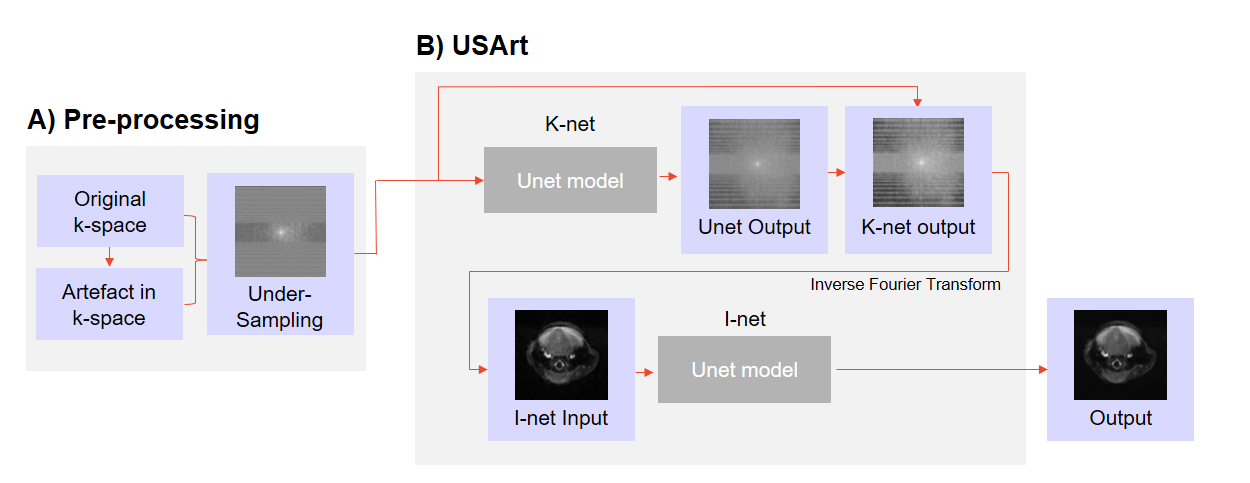}
  \caption{The preprocessing pipeline and USArt. A) Artifacts and noise are added to full k-space, before under-sampling is performed using specific masks and acceleration factors. This degraded k-space dataset is used as input for the USArt model. B) USArt utilizes two U-Net based components: K-net and I-net. K-net operates in the k-space domain to fill missing lines, and its output is transformed to the image domain via an inverse Fourier Transform. I-net then refines this output, focusing on artifacts correction and image consistency.}
    \label{fig:methodsmodel} 
\end{figure}

\subsubsection{Gaussian Noise}
The noise present in MRI usually originates from electronic sources involved in the data acquisition process. In the k-space domain, Gaussian noise is typically observed, resulting in Rician noise in the reconstructed images \cite{Constantinides1997,Dietrich2007}. In this study, the noise was applied in the k-space as an additive complex Gaussian random signal, i.e. with the real and imaginary parts being independent random variables with zero mean, and variance set according to the simulated SNR. The maximum noise level caused a reduction of the original image SNR by half. 

\subsubsection{Motion artefacts}

Motion artefacts (MA) in MRI typically originate from patient motion; this motion can lead to inconsistencies in the magnetic field interactions, resulting in deviations from the expected signal patterns. MA primarily affect the acquired data (k-space data) and manifest as discrepancies or distortions in the k-space lines corresponding to the moment of motion. The appearance of MA depend on the k-space scanning strategy, particularly whether the acquisition is conducted in 2D or 3D. Movements that occur when the acquisition trajectory corresponds to low frequencies (central k-space lines) often result in ghosting artefacts, while movements occurring during high-frequency acquisitions often result in ringing and blurring artefacts. 
In this work, MA were implemented as rotation or shift in the image domain as described by~\cite{Shaw2020} and its effect on k-space data reflected by the Fourier Transformation of the artefacted images.

\subsection{Metrics}
To assess the quality of the images, we used the well-established reference-based Structural SIMilarity (SSIM) index focused on the subject only by removing background (SSIMf)~\cite{Qiu2020} and Peak Signal-to-Noise Ratio (PSNR). These metrics allow an objective comparison between USArt reconstructed images and the ground truth images. Since ground truth images themselves are not perfect and always contain some level of degradation, we also used reference-free metrics, including SNR and Contrast to evaluate an absolute image quality on an individual basis. Contrast was measured as the standard deviation of tissue signal.

\begin{table*}[h]
\centering
\small
    \caption{Performance of our proposed model with various under-sampling strategies, acceleration factors, and artefacts. The first three lines compare USArt performance with different under-sampling strategies. The next three lines show USArt performance with different acceleration factors. The next 4 lines show the robustness of USArt to the presence of artefacts. The bottom part of the table provides benchmark values for the reference KIKI~\cite{Eo2018} model using a $5\times$ acceleration factor and gradient under-sampling in the presence of artifacts, showing the superiority of our model in real-world acquisitions.}
      
    \scalebox{0.85}{
    \begin{tabular}{llc||cccc}
    \label{tab:resultsT}
        Acc. & US masks & Artifact  &  SSIMf & PSNR  & SNR  & Contrast  \\
        \hline
        & Original & No &  & & 15.249 $\pm$ 2.170  & 0.671 $\pm$ 0.141    \\
        \hline
        \hline
        
        \multicolumn{6}{l}{} \\
        \multicolumn{2}{l}{\textbf{USArt}} \\
        5x & Gradient & No &\textbf{ 0.971 $\pm$ 0.008} & \textbf{ 77.219 $\pm$ 1.436}  &  \textbf{ 78.976 $\pm$ 11.125} & 0.710 $\pm$ 0.148  \\ 
        5x &  Random  & No & 0.969 $\pm$ 0.009 & 76.586 $\pm$ 1.467 & 78.946 $\pm$ 11.868 & 0.709 $\pm$ 0.148 \\
        5x & Uniform & No & 0.969 $\pm$ 0.009 &  76.541 $\pm$  1.541 & 75.712 $\pm$ 11.558 & \textbf{0.721 $\pm$ 0.152}  \\
        \hline
        
        2x & Gradient & No &  0.979 $\pm$ 0.006 & 78.477 $\pm$ 1.457 &  51.656 $\pm$ 4.290 & 0.708 $\pm$ 0.137  \\
        5x &  Gradient & No &  0.971 $\pm$ 0.008 & 77.219 $\pm$ 1.436 &  75.712 $\pm$ 11.558 & 0.710 $\pm$ 0.148\\  
        10x & Gradient & No &  0.962 $\pm$ 0.011 & 74.934 $\pm$ 1.798 &  107.222 $\pm$ 16.006 & 0.746 $\pm$ 0.148 \\
        \hline
        
        5x &  Gradient & No &  0.971 $\pm$ 0.008 & \textbf{77.219 $\pm$ 1.436} &  \textbf{75.712 $\pm$ 11.558} & 0.710 $\pm$ 0.148\\ 
        5x & Gradient & Noise &  \textbf{0.966 $\pm$ 0.010} & \textbf{76.126 $\pm$ 1.747} & \textbf{85.638 $\pm$ 14.070} & \textbf{0.735 $\pm$ 0.149}  \\
        5x & Gradient & MA &\textbf{ 0.964 $\pm$ 0.011 }& \textbf{75.141 $\pm$ 1.816} & 55.759 $\pm$ 08.380 & \textbf{0.705 $\pm$ 0.143}  \\
        5x & Gradient  & N+MA & \textbf{0.960 $\pm$ 0.013} & \textbf{74.557 $\pm$ 1.607} & \textbf{84.251 $\pm$ 16.347} & \textbf{0.743 $\pm$ 0.143} \\
             
        \hline
        \hline
        
        \multicolumn{6}{l}{} \\
        \multicolumn{2}{l}{\textbf{KIKI~\cite{Eo2018}}} & & & &\\
        5x & Gradient & No  & \textbf{0.971 $\pm$ 0.007} & 76.496 $\pm$ 1.175 & \textbf{77.125 $\pm$ 11.824} & \textbf{0.735 $\pm$ 0.154} \\
        5x & Gradient & Noise  & 0.954 $\pm$ 0.049 & 76.091 $\pm$ 1.303 & 44.297 $\pm$ 19.010 & 0.677 $\pm$ 0.140 \\
        5x & Gradient & MA & 0.952 $\pm$ 0.013 & 74.935 $\pm$ 1.952 & \textbf{57.293 $\pm$ 20.947} & 0.642 $\pm$ 0.162 \\
        5x & Gradient &  N+MA & 0.951 $\pm$ 0.017 & 73.179 $\pm$ 3.408 & 40.172 $\pm$ 19.683 & 0.612 $\pm$ 0.164 \\
        \hline
        \hline
             
    \end{tabular}
    }
\end{table*}

\subsection{Pre-processing}
The pre-processing pipeline (Fig.~\ref{fig:methodsmodel}-A) begins by introducing artefacts, such as MA and/or noise, into the k-space domain. Subsequently, the data experience under-sampling, a process that zeros specific lines in the k-space. To achieve this, one of the three under-sampling masks and acceleration factors is applied. These steps collectively result in a k-space dataset degraded by both artefacts and under-sampling, which acts as the input for our USArt.

\begin{figure}[h]
  \centering
  \includegraphics[width=0.86\textwidth]{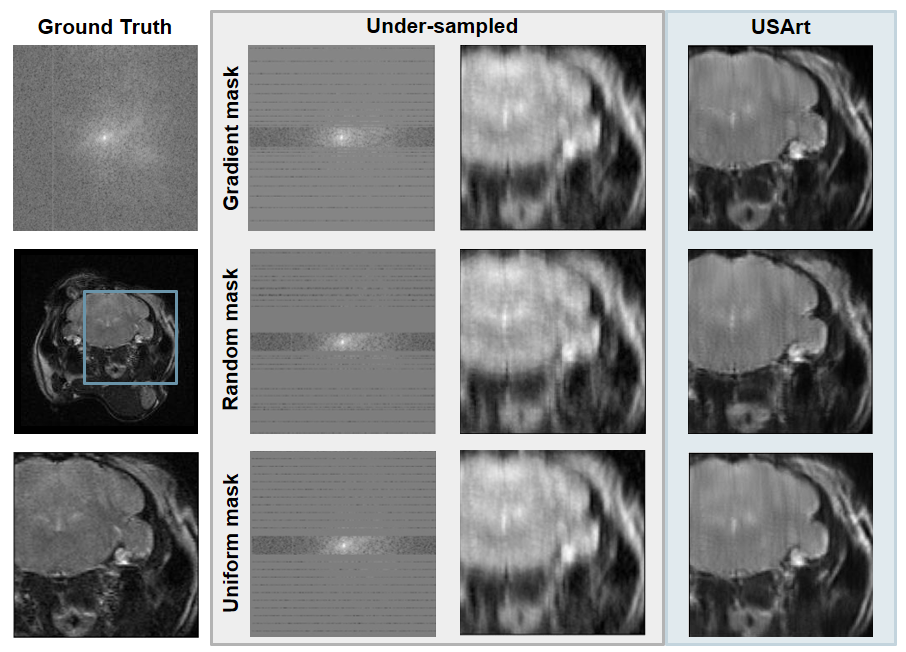}
  \caption{Under-sampling strategies with acceleration factor $5\times$: Left column from top to bottom) k-space, reconstructed, and zoom images (blue frame) for the original image. Grey panel) Under-sampled k-space and the zoom corresponding reconstructed images for gradient, random and uniform under-sampling. Blue-light panel) the corresponding USArt's output.}
    \label{fig:resultsmask} 
\end{figure}

\begin{figure}[h]
  \centering
  \includegraphics[width=0.86\textwidth]{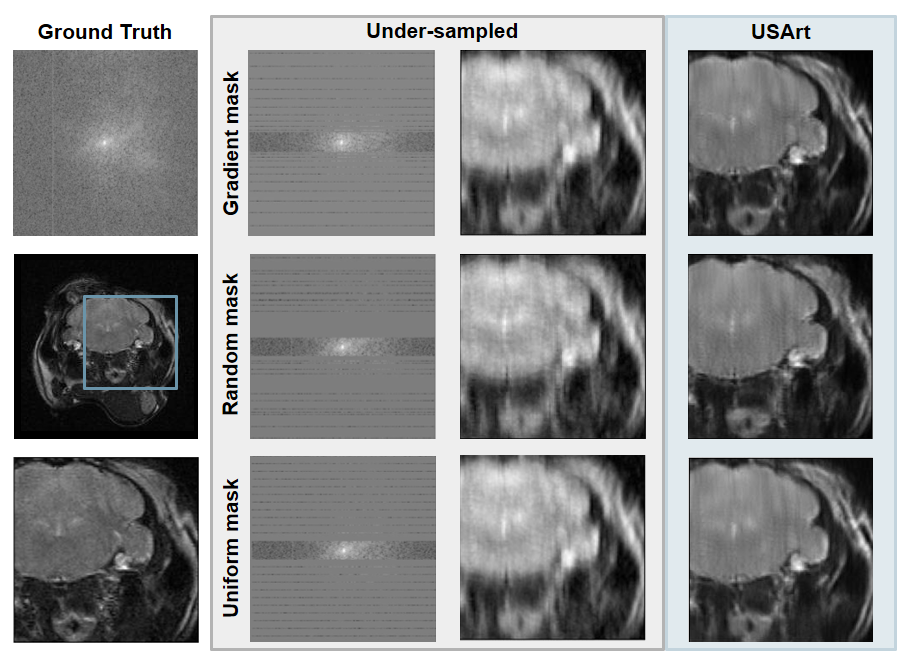}
  \caption{Different acceleration factors: Left column) from top to bottom: k-space, reconstructed, and zoom images (blue frame) from the original image. Grey panel) Under-sampled k-space and the zoom corresponding reconstructed images for acceleration factors $2\times$, $5\times$, and $10\times$ . Blue panel) the corresponding USArt's output.}
    \label{fig:resultsmodel} 
\end{figure}

\subsection{Proposed model}

We address the correction of image degradation caused by under-sampling in the presence of artefacts by using a dual model strategy, that has recently showed successes in the context of MR images~\cite{Haji2021,Eo2018,Cheng2019}. Considering the complex nature of the MR k-space, we divide MR complex data into real and imaginary parts and use them as two input channels in the model. In USArt, 
the first part (K-net) focuses solely on the under-sampled k-space, while the second part (I-net) concentrates on improving image quality and consistency  (Fig.~\ref{fig:methodsmodel}-B). Both parts (K-net and I-net) are based on the U-Net architecture; U-Net was proposed for biomedical image segmentation~\cite{AnbuDevi2022} and has since become a popular choice also for image-to-image translation tasks. 
In detail, K-net operates within the k-space domain, addressing the task of filling the missing lines. The output of the K-net is modified by inserting back the acquired k-space lines while keeping the lines predicted by the K-net model at the k-space positions with no data acquired. Transformation from the k-space domain to the image domain is achieved by an inverse Fourier Transform operation. In the last step, I-net operates within the image domain, focusing on artefacts correction and image consistency improvement. The K-net and I-net networks were trained sequentially; after K-net training was completed, I-net was trained using the output from K-net. We compare USArt\footnote{https://github.com/TransRad/USArt.git} with the KIKI model~\cite{Eo2018}; KIKI has focused on utilizing strategy for reconstructing under-sampled image data, our work extends its application to concurrently tackle both under-sampling and artefacts. 

The image input size is 2x256x256. U-Net with 4 layers is trained for 150 epochs with batch size 16 and an initial learning rate of 0.0001. We use AdamW optimization as a stochastic gradient descent method. A learning rate decay with 20 epochs of patience is applied with 10\% drop. For the activation function, a leaky rectified linear unit (Leaky-ReLU) is used with negative slope 0.2. L2 regularization penalties are applied on a per-layer basis. We used a focus Multi-Scale Structural Similarity (MS-SSIM)~\cite{Wang2003} loss function of the missing lines of the k-space for the K-net network and MS-SSIM of the whole reconstructed image for the I-net network in order to minimize the loss error. The networks are implemented using the TensorFlow framework. The training and inference used a GPU NVIDIA GeForce RTX 3080Ti graphics card with 12GB RAM.

\begin{figure}[h]
  \centering
  \includegraphics[width=0.9\textwidth]{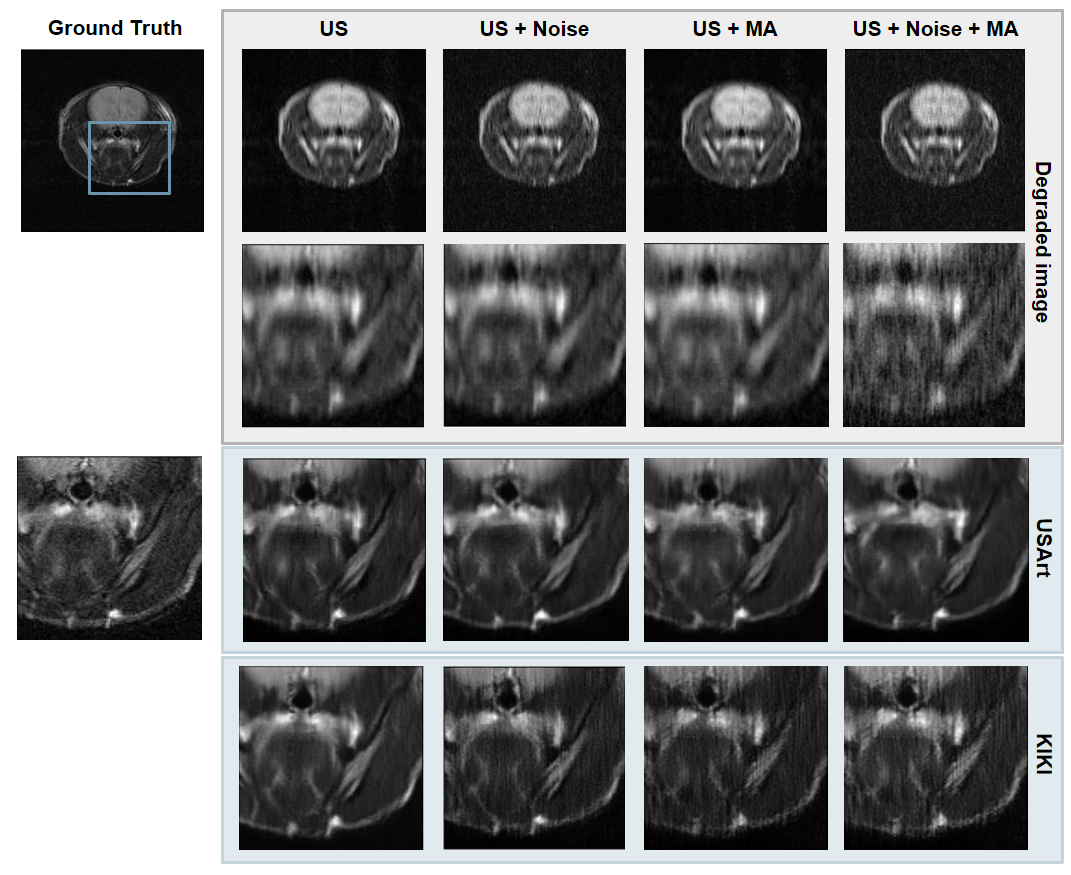}
  \caption{Illustration of artefacts correction in accelerated images: First column) Original and zoom images (blue frame). Grey box) $5\times$under-sampled full and zoomed images with no artefacts, noise, motion artefact and their combination. Blue box) Corresponding images with quality restored by USArt and KIKI model's. US: Under-sampled, MA: Motion Artifact.}
    \label{fig:resultsAR} 
\end{figure}

\section{Results}

In preparation for combining the acceleration and artefacts correction, we first established the optimal under-sampling (US) strategy and the acceleration factor. The MRI details showcased in Fig.~\ref{fig:resultsmask} present variations under different under-sampling strategies. The highest SSIMf and PSNR were achieved with the gradient under-sampling mask. For the reference free metrics, the gradient under-sampling provided the best results in SNR, while the uniform under-sampling mask provided the best result for Contrast (Tab.~\ref{tab:resultsT}). The gradient under-sampling strategy was thus retained for the subsequent tasks. 

We then evaluated USArt using various acceleration factors, and were able to efficiently reconstruct quality images for all cases (Fig.~\ref{fig:resultsmodel}).
We found that $5\times$ acceleration was possible without significant degradation (Tab.~\ref{tab:resultsT}), with SSIMf to 0.971 and PSNR to 77.219. Subsequently, the acceleration factor $5\times$ was retained for the ensuing tests.

Using the gradient under-sampling strategy and an acceleration factor of $5\times$ we then evaluated the simultaneous correction of under-sampled data and artefacts, with the model detailed earlier. USArt effectively restored quality in under-sampled acquisitions, despite the presence of artefacts (Fig.~\ref{fig:resultsAR}). The results were confirmed by the metrics in Tab.~\ref{tab:resultsT}, where only a limited reduction was observed in the reconstructed under-sampled images in the presence of artefacts, compared to those obtained in the absence of artefacts, with reductions of less than 1.2\% in SSIM and 3.5\% in PSNR, respectively. The results from USArt outperformed those from previous well-established models used for image restoration from under-sampled data, notably the KIKI model~\cite{Eo2018}, establishing the superiority of our model in real-word settings. 

\section{Discussion and Perspectives}
Acquiring high-quality MR images can be time-consuming. Reducing k-space sampling saves time but typically lowers image quality. The USArt model introduces a novel approach that simultaneously improves image quality and corrects artifacts in accelerated MR imaging. The K-net and I-net sub-models work together to enhance image details, restore contrast, and ensure image consistency. Moreover, the model demonstrates robustness against real-world degradations such as noise and motion artifacts, even when applied to under-sampled data.

Having established confidence in the model's capacity to manage simultaneous under-sampling and artefact correction, future research could explore alternative model architectures such as Vision Transformers. These architectures, proven effective in image-to-image translation tasks, hold promise for enhancing our USArt. Moreover, our approach is also applicable to accelerating clinical data and other types of trajectories and more advanced protocols, where various types of artefacts may be observed.

\bibliographystyle{ieeetr}
\small
\bibliography{bib}

\end{document}